%% file: iclr2025_conference.tex
\documentclass{article} % For LaTeX2e
\usepackage{iclr2025_conference,times}

\input{math_commands.tex}

\usepackage{hyperref}
\usepackage{url}
\usepackage{multirow}
\usepackage{arydshln}  % 引入虚线包
\usepackage{booktabs}
\usepackage{graphicx}

\title{FluentLip: A Phonemes-Based Two-stage Approach for Audio-Driven Lip Synthesis with Optical Flow Consistency}

\author{Shiyan Liu, Rui Qu \& Yan Jin \\
Department of Computer Science\\
Huazhong University of Science and Technology\\
Hubei, China \\
\texttt{\{shyl,rqu,jinyan\}@hust.edu.cn} \\
}
\begin{document}

\maketitle

\begin{abstract}
Generating consecutive images of lip movements that align with a given speech in audio-driven lip synthesis is a challenging task. While previous studies have made strides in synchronization and visual quality, lip intelligibility and video fluency remain persistent challenges. This work proposes FluentLip, a two-stage approach for audio-driven lip synthesis, incorporating three featured strategies. To improve lip synchronization and intelligibility, we integrate a ref encoder to generate a fusion of audio and phoneme information for multimodal learning. Additionally, we employ optical flow consistency loss to ensure natural transitions between image frames. Furthermore, we incorporate a diffusion chain during the training of Generative Adversarial Networks (GANs) to improve both stability and efficiency. We evaluate our proposed FluentLip through extensive experiments, comparing it with five state-of-the-art (SOTA) approaches across five metrics, including a proposed metric called Phoneme Error Rate (PER) that evaluates lip pose intelligibility and video fluency. The experimental results demonstrate that our FluentLip approach is highly competitive, achieving significant improvements in smoothness and naturalness. In particular, it outperforms these SOTA approaches by approximately $\textbf{16.3\%}$ in Fréchet Inception Distance (FID) and $\textbf{35.2\%}$ in PER.

\end{abstract}

\section{Introduction}
Audio-driven lip synthesis, also known as Talking Face Generation (TFG), generates a coherent sequence of mouth movements that are consistent with the given audio input. It has become a prominent topic of research ~\citep{jamaludin2019you} due to its wide range of real-world applications, such as film dubbing~\citep{kim2018deep}, video bandwidth reduction~\citep{suwajanakorn2017synthesizing} and face animation~\citep{song2019talking}. Despite its potential, achieving perfect lip synchronization remains a significant challenge. Hence, it has attracted considerable attention from researchers.

Numerous methods have been proposed to improve synchronization and visual quality in audio-driven lip synthesis. To enhance synchronization, Wav2Lip~\citep{prajwal2020lip} extends SyncNet~\citep{chung2017out} to the RGB space, using a lip sync discriminator to calculate sync loss and penalize asynchronous lip pose generated. SyncTalkface~\citep{park2022synctalkface} calculates sync loss by measuring lip pose feature distances between synthesized and ground truth videos. TalkLip~\citep{wang2023seeing} leverages a pre-trained lip-reading expert~\citep{shilearning} to guide lip pose synthesis. Moreover, many approaches, like Wav2Lip, incorporate Generative Adversarial Networks (GANs)~\citep{goodfellow2014generative} to enhance visual quality.

While synchronization and visual quality have been well-studied, less attention has been given to improving lip pose intelligibility and video fluency. Notably, TalkLip employs a lip-reading expert to enhance lip pose intelligibility~\citep{wang2023seeing}. In our work, we address these gaps by proposing a phoneme-based two-stage approach with optical flow consistency (denoted as FluentLip), specifically designed to improve both lip pose intelligibility and video fluency.

Specifically, we utilize a phoneme extractor to automatically recognize and align phonemes from the audio. The ref encoder then transforms audio and phonemes into embeddings, serving as the reference input for the lip sync discriminator and generator within GANs. To further enhance fluency, we introduce an optical flow consistency loss that penalizes unnatural transitions between frames during training. Additionally, we employ a diffusion model~\citep{ho2020denoising} with a diffusion chain to accelerate convergence and stabilize the training process~\citep{wang2023diffusiongan}, ultimately improving visual quality. 

% Specifically, we utilize a phoneme extractor to automatically recognize and align phonemes from the audio and a phoneme encoder to generate the corresponding phoneme embeddings. These embeddings are then fused with audio embeddings to serve as the reference input for the lip sync discriminator and generator within GANs. To further enhance fluency, we introduce an optical flow consistency loss that penalizes unnatural transitions between frames during training. Additionally, we employ a diffusion model~\citep{ho2020denoising} with a diffusion chain to accelerate convergence and stabilize the training process~\citep{wang2023diffusiongan}, ultimately improving visual quality. 

The main contributions of this work are summarized as follows.
\begin{itemize}
    \item We leverage a ref encoder to create a fusion of phoneme and audio embeddings, improving lip pose intelligibility. Additionally, we develop an optical flow consistency loss to guide training, ensuring smooth transitions between frames and enhancing the naturalness of synthesized videos. Besides, we employ a diffusion chain to enhance the stability of the training process. Our approach specifically addresses the underexplored challenges of lip pose intelligibility and video fluency in audio-driven lip synthesis.
    % \item We leverage a phoneme extractor and encoder to create a fusion of phoneme and audio embeddings, improving lip pose intelligibility. Additionally, we develop an optical flow consistency loss to guide training, ensuring smooth transitions between frames and enhancing the naturalness of synthesized videos. Our approach specifically addresses the underexplored challenges of lip pose intelligibility and video fluency in audio-driven lip synthesis.
    % \item We integrate a diffusion chain into the training process of GANs, leading to faster convergence and enhancing the stability of the training process. The quality of the synthesized videos is improved, providing realistic and visually appealing outputs that align closely with the corresponding audio input.
    \item We evaluate the effectiveness of our proposed FluentLip with five state-of-the-art (SOTA) approaches, demonstrating a notable performance of approximately 16.3\% in Fréchet Inception Distance (FID) and 35.2\% in Phoneme Error Rate (PER). Additionally, we introduce a novel metric that leverages insights from the lip-reading expert and the Grapheme-to-Phoneme (G2P) model to assess the perceptual performance of various approaches.
\end{itemize}

\section{Related Work}

\subsection{Speech-Driven Talking Face Generation}

Talking face generation was first proposed in the 1990s~\citep{yehia1998quantitative}, with early approaches primarily using Hidden Markov Models (HMM)~\citep{bregler1997video}. In recent years, deep learning has emerged as the dominant TFG method, which can be generally classified into intermediate representation-based approaches and reconstruction-based approaches~\citep{park2022synctalkface}. 

The intermediate representation-based approaches focus on learning facial representations, such as 3D meshes, which are used for facial synthesis. For example, SadTalker~\citep{zhang2023sadtalker} generates 3D-aware face renders for synthesizing talking faces, while Everybody’s Talkin’~\citep{song2022everybody} reconstructs 3D meshes from extracted facial parameters to generate video sequences. However, these approaches are limited in their generalizability to arbitrary characters, and 3D modeling often struggles to represent mouth details~\citep{wang2023seeing}. 

In contrast, reconstruction-based approaches primarily rely on end-to-end encoder-decoder architectures, which avoid the limitations of intermediate representations and offer improved mouth details synthesis. It began with ObamaNet~\citep{kumar2017obamanet}, which focused on a specific character. This was followed by Speech2Vid\citep{jamaludin2019you} and LipGAN~\citep{kr2019towards}, which improved generalizability allowing for video generation of arbitrary characters. A breakthrough came with Wav2Lip \citep{prajwal2020lip}, which introduced SyncNet~\citep{chung2017out} as a lip sync expert, achieving SOTA synchronization performance. More recent efforts have aimed at improving visual quality based on the Wav2Lip model. \citet{gupta2023towards} pre-trained a VQGAN model~\citep{esser2021taming} to train Wav2Lip in quantized space, improving visual quality to a maximum of 4K resolution. Diff2Lip~\citep{mukhopadhyay2024diff2lip} uses a diffusion model~\citep{ho2020denoising} to replace the Seq2Seq framework in Wav2Lip, further improving visual quality. 
%For instance, Wav2Lip~\citep{prajwal2020lip} employs a lip sync expert to enhance synchronization, while Diff2Lip~\citep{mukhopadhyay2024diff2lip} leverages a diffusion model-based generator to efficiently improve visual quality.

%The reconstruction-based TFG algorithms have more variants, which can be divided into two categories: some algorithms focus on basic synchronization and visual quality; other algorithms are based on the former algorithms, focusing on features that are more concerned with the realism and watching experience. The former starts with ObamaNet~\citep{kumar2017obamanet}, working on just one specific character. Subsequently, Speech2Vid\citep{jamaludin2019you} and LipGAN~\citep{kr2019towards} are proposed to improve the generalizability and can generate videos for arbitrary characters. After that, \citet{prajwal2020lip} propose Wav2Lip, introducing SyncNet~\citep{chung2017out} as a Lip Sync Expert, reaching peak performance on synchronization. Since then, lots of work tend to improve the visual quality. \citet{gupta2023towards} pre-train a VQGAN model~\citep{esser2021taming} to train Wav2Lip in quantized space, improving the visual quality to a maximum of 4K. Diff2Lip~\citep{mukhopadhyay2024diff2lip} uses a diffusion model~\citep{ho2020denoising} to replace the Seq2Seq framework in Wav2Lip, which also improves the visual quality. 

At the same time, some works have taken an alternative approach by focusing on issues that impact human viewing, particularly by integrating generated videos into real-life scenarios. Among the most novel concerns is lip pose intelligibility. \citet{wang2023seeing} is the first to highlight this issue in the context of TFG, introducing AV-HuBERT~\citep{shilearning} as a lip-reading expert to improve lip pose intelligibility and opening a new direction for TFG research. 

Despite the increasing number of works in TFG, surprisingly little attention has been given to video fluency. Although these subtle details may be difficult to perceive with the naked eye, as visual quality continues to improve, fluency will become a critical issue. To fill this gap, our work introduces phonemes, commonly used in the Text-to-Speech (TTS) domain, along with a novel metric to promote assessing lip pose intelligibility. Furthermore, we incorporate optical flow consistency loss to improve the fluency of generated videos.

%Hence, our work introduces phonemes, which are more prevalent in the Text-to-Speech (TTS) domain, and a novel metric to enhance lip pose intelligibility and measure it more precisely. What’s more, we introduce optical flow consistency loss to improve the fluency of the generated videos.

\subsection{Phoneme-Based Multimodal Learning}

Most previous TFG studies employ audio or text as the input driver with their unimodal learning model. ATVGnet~\citep{chen2019hierarchical} and Wav2Lip~\citep{prajwal2020lip} employ audio as driven input, while ParaLip~\citep{liu2022parallel} and Make-A-Video~\citep{singermake} are text-driven. Since audio varies with different speakers and text may contain homophones, it's difficult to represent speech content with just one of them. Phoneme~\citep{zhang2022text2video} is a more microscopic concept widely used in the TTS domain, focusing on syllables rather than words. Although some previous works employ phonemes in TFG, such as Text2video~\citep{zhang2022text2video} and text-based editing video~\citep{fried2019text}, few of them have extended unimodal to multimodal learning. Thus, we introduce multimodal learning in our work by combining audio and phoneme as driven inputs. Phonemes capture precise speech content, while audio conveys robust and ample information, helping to judge speech content accurately and ultimately enhancing lip pose intelligibility.

\subsection{Consecutive Image Generation}

Video generation can be viewed as a process of generating consecutive images in frames, and the three primary frameworks are prevalent for solving it: Seq2Seq, GANs~\citep{goodfellow2014generative}, and diffusion model~\citep{ho2020denoising}. Among these, Seq2Seq serves as the foundational model, while diffusion models have demonstrated outperforming GANs in image generation~\citep{dhariwal2021diffusion}. Most of the previous TFG works use Seq2Seq to generate frame images, often coupled with GANs to enhance the visual quality of these images, such as Wav2Lip~\citep{prajwal2020lip}. Some approaches use the diffusion model as an image generator, which also achieves good results, such as Diff2Lip~\citep{mukhopadhyay2024diff2lip}. Nevertheless, GANs training often comes up with the mode collapse~\citep{wang2023diffusiongan}, a challenge that has been overlooked in previous TFG works that leverage GANs. To mitigate this, \citet{wang2023diffusiongan} propose Diffusion-GAN, which integrates a diffusion model into the GANs training to generate Gaussian instance noises in high-dimensional data space, effectively improving the stability and overall performance of GANs training. Inspired by it, our work also employs a diffusion model to stabilize GANs training and improve its performance.

Unlike naive image generation, consecutive image generation should consider the fluency between frames. In real-life videos, objects are moving regularly with specific trends, so the pixel points move more smoothly. Naive image generation methods often neglect this, leading to irregular and trendless pixel point movement between frames. Optical flow~\citep{horn1981determining}, a technique frequently used to measure pixel displacement between two consecutive frames, is estimated by FlowNet~\citep{dosovitskiy2015flownet, ilg2017flownet} or more popular Recurrent All-pairs Field Transforms (RAFT)~\citep{teed2020raft} and is often applied in dynamic image detection. For example, self-driving automobiles use optical flow to predict the motions and traces of surrounding objects~\citep{hu2020multi}. Therefore, we assert that optical flow is an effective measure of video fluency and design a novel loss function based on optical flow to penalize irregular pixel moves generated, enhancing video fluency. Note that our work first employs optical flow in the TFG domain.

\section{The Proposed FluentLip Approach}
The two-stage approach that combines a lip sync discriminator and a lip synthesis network has proved to be quite successful for audio-driven lip synthesis~\citep{prajwal2020lip, gupta2023towards, mukhopadhyay2024diff2lip}. Leveraging this powerful framework, we design dedicated fused embedding and optical flow consistency strategies to address lip pose intelligibility and video fluency, and to improve lip synchronization. 

\begin{figure}
    \centering
    \includegraphics[width=\linewidth]{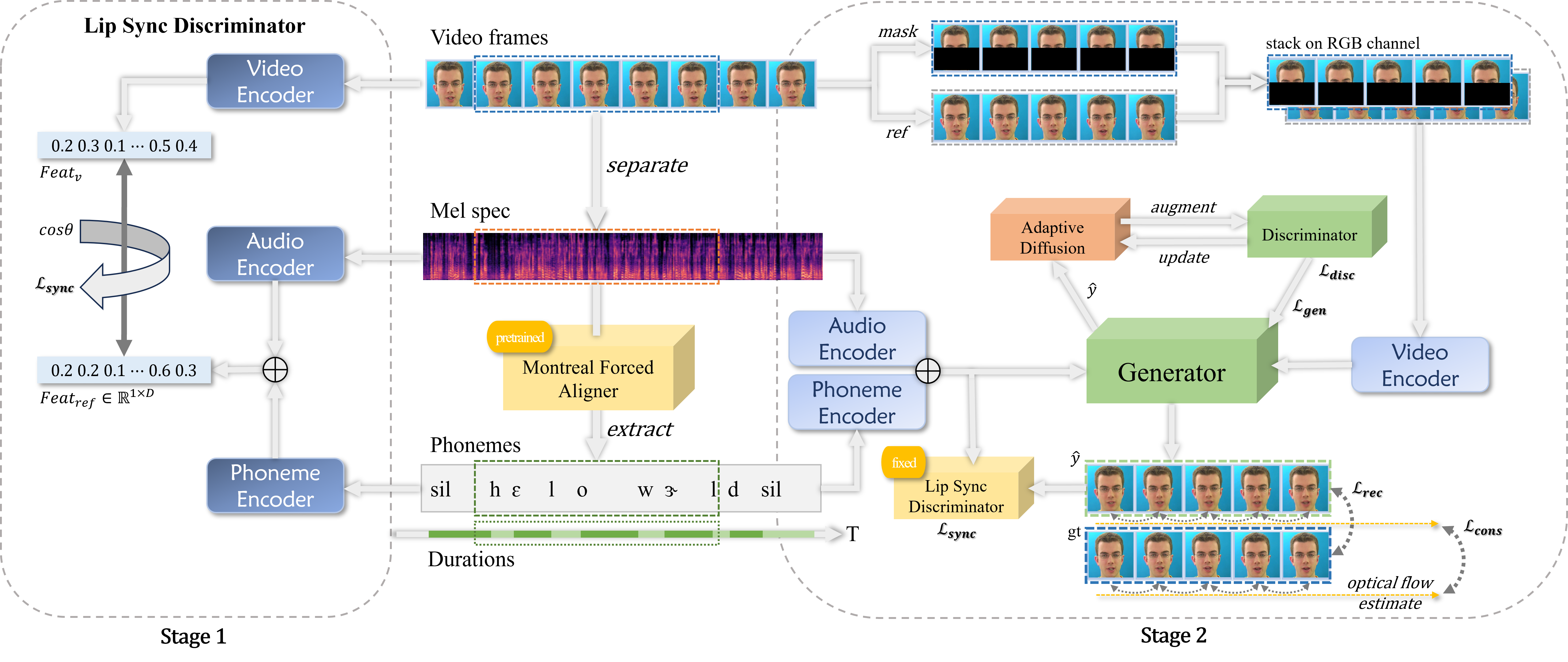}
    \caption{The architecture of the proposed FluentLip approach}
    \label{FigStage1}
\end{figure}

Algorithm \ref{FigStage1} outlines the architecture of FluentLip, which adopts a two-stage process. In stage 1, phonemes are automatically extracted from the audio corresponding to a given video frame, and aligned precisely by frame. A ref encoder, containing an audio encoder and a phoneme encoder, generates fused audio and phoneme embeddings, which are then fed into the lip sync discriminator alongside the corresponding video embeddings. This fusion of sensory modalities establishes multimodal learning.

% Algorithm \ref{FigStage1} outlines the architecture of FluentLip, which adopts a two-stage process. In stage 1, phonemes are automatically extracted from the audio corresponding to a given video frame, and aligned precisely by frame. A phoneme encoder generates phoneme embeddings, which are then fused with the audio embeddings and fed into the lip sync discriminator alongside the corresponding video embeddings. This fusion of sensory modalities establishes multimodal learning.

In stage 2, the fused audio and phoneme embeddings are used to train the lip generator, together with video embeddings from stacked frames of both predicted and reference images. The synthesized facial video is guided by the fixed lip sync discriminator from stage 1 via a sync loss to ensure precise lip synchronization, and by the visual discriminator of GANs to improve visual quality. 

Additionally, an adaptive diffusion model is employed between the generator and visual discriminator, where a diffusion chain of variable length is applied to gradient propagation to improve the stability and effectiveness of the training process. To further improve the realism of the synthesized video, the RAFT model predicts optical flow between frames, applying optical flow consistency loss to penalize unnatural shifts. All losses are integrated to optimize the training of the whole network. Below, we provide a detailed description of each core component of the FluentLip approach.

\subsection{Stage 1: Lip Sync Discriminator}
\label{SecStage1}

\textbf{Phoneme encoder} The phoneme encoder is a part of the ref encoder mentioned above to effectively encode the phoneme sequence, which is subsequently concatenated and fused with the audio embedding. The raw phoneme text and its corresponding durations are automatically extracted from the reference audio by a pre-trained phoneme extractor Montreal Forced Aligner (MFA)~\citep{mcauliffe2017montreal}, as illustrated in Fig. \ref{FigStage1}. The phoneme text sequence is first converted into numerical representations via a global phoneme table. Given that the length of the phoneme sequences varies across different audio clips, we pad both the phoneme encodings and their corresponding duration vector to a fixed length before proceeding with the embedding process. Within the phoneme encoder, positional encoding is employed alongside a Transformer-based architecture, which improves the model’s ability to capture the sequential dependencies inherent in the phonemes, ultimately generating high-quality phoneme embeddings.

%The phoneme encoder is adopted in each stage to encode the phoneme sequence, which is then concatenated and fused with the audio embedding. The raw phoneme text and its corresponding duration are automatically extracted from the reference audio by a pre-trained phoneme extractor called Montreal Forced Aligner (MFA)~\citep{mcauliffe2017montreal} in Fig. \ref{FigStage1}. The phoneme text sequence is firstly encoded into numbers via a global phoneme table. Considering that the lengths of phoneme sequence of each audio clip may not be consistent, we pad the phoneme encoding and duration vector to a fixed length before embedding. Inside the phoneme encoder, a positional encoding and a Transformer-based network are used, which is helpful to capture the sequential information behind the phonemes better, to produce the phoneme embedding.

\begin{figure}
    \centering
    \includegraphics[width=1\linewidth]{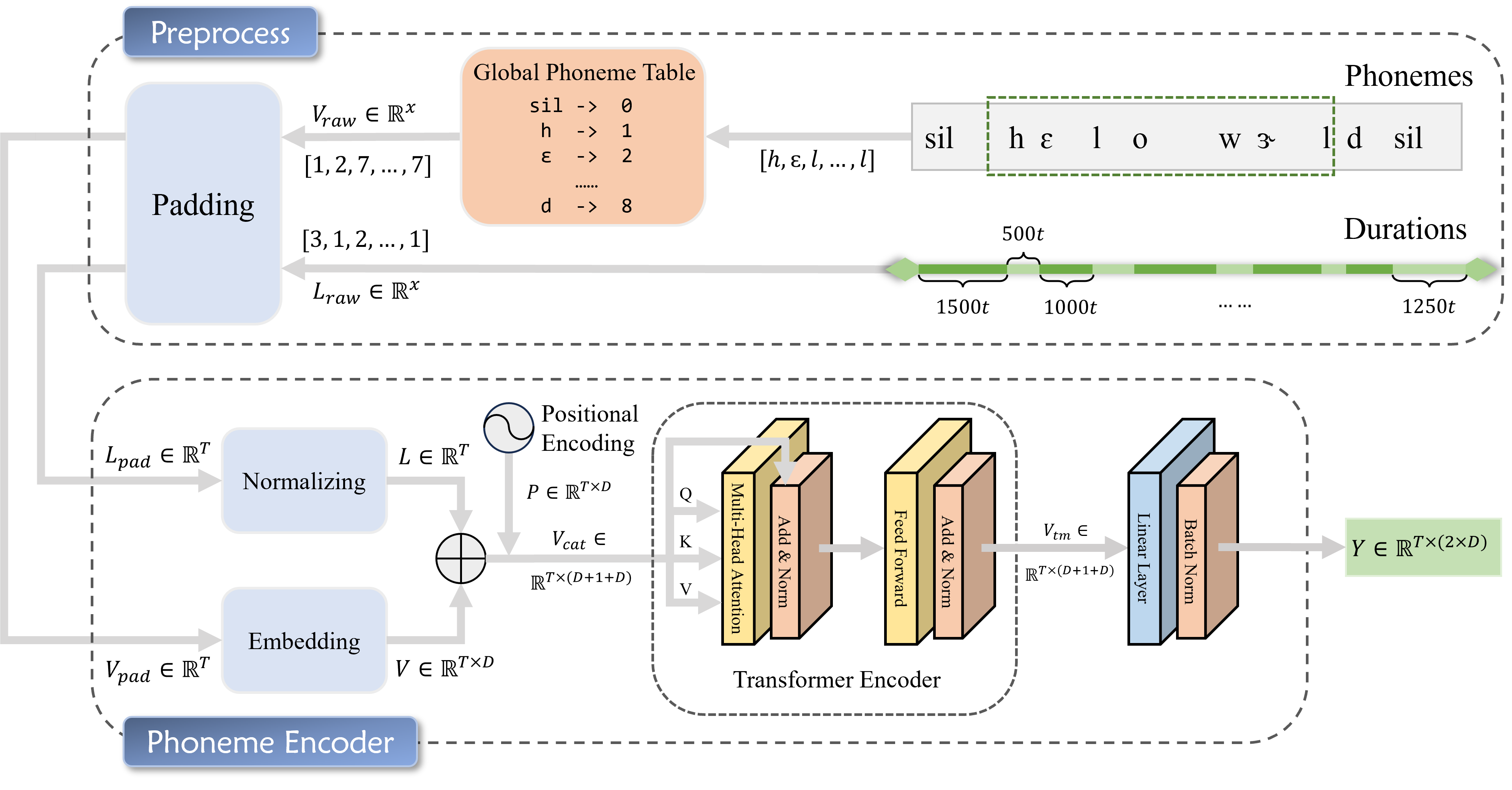}
    \caption{The phoneme encoding procedure, with $\mathit{t}$ representing the unit time for duration.}
    \label{FigEncoder}
\end{figure}

Let us denote $\mathit{V_{raw}}\in\mathbb{R}^{x}$ and $\mathit{L_{raw}}\in\mathbb{R}^{x}$ as the initial phoneme encoding and duration vector respectively, where $\mathit{x}$ is the irregular length of each phoneme vector. The padded phoneme encoding and duration vector are denoted as $\mathit{V_{pad}}\in\mathbb{R}^{T}$ and $\mathit{L_{pad}}\in\mathbb{R}^{T}$, where $\mathit{T}$ is the fixed phoneme vector length. The embedded phoneme vector is represented as $\mathit{V}\in\mathbb{R}^{T\times{D}}$, where $\mathit{D}$ is the feature dimension, and $\mathit{L}\in\mathbb{R}^{T}$ is the normalized duration vector. Additionally, $\mathit{P}\in\mathbb{R}^{T\times{D}}$ denotes the positional encoding vector. The sequential concatenation of $\mathit{V}$, $\mathit{L}$ and $\mathit{P}$, denoted as $\mathit{V_{cat}\in\mathbb{R}^{T\times(D+1+D)}}$, is fed to the Transformer network, which processes the input to generate the penultimate phoneme embedding $\mathit{V_{tm}\in\mathbb{R}^{T\times(D+1+D)}}$. Subsequently, a linear layer followed by a batch normalization layer produces the ultimate phoneme embedding $\mathit{Y\in\mathbb{R}^{T\times(D\times2)}}$, which serves as the output. The whole procedure of phoneme encoding is shown in Fig. \ref{FigEncoder}.

\textbf{Lip sync discriminator} The lip sync discriminator, such as the previously proposed SyncNet~\citep{chung2017out}, aims to evaluate the synchronization between a Mel spectrum clip and a lip motion clip by comparing their embeddings under a latent space. Inspired by Wav2Lip~\citep{prajwal2020lip}, which first applied the lip sync discriminator into the TFG to improve lip synchronization, we also integrate this module, introducing phonemes as a novel addition. The cosine similarity between the audio and video feature vectors is calculated and used to obtain the sync loss by computing Binary Cross-Entropy (BCE). 

Let us denote $\mathit{y}$ as the target similarity, whose value reflects whether the audio-video pair is originally matched, and $\mathit{S(m, n)}$ is the cosine similarity function for feature vectors $\mathit{m}$ and $\mathit{n}$. For $\mathit{N_i}$ audio-video pairs with audio embedding $\mathit{a}$ and video embedding $\mathit{v}$, the sync loss is formulated as:
\begin{equation}
    \mathcal{L}_{sync} = \frac{1}{N_i}\sum_{i}^{N_i}[\textnormal{-}y_i\log{S(a_i, v_i)} - (1-y_i)\log{(1-S(a_i, v_i))}]
\end{equation}
We fuse the phoneme embedding, extracted from the audio and encoded by the phoneme encoder, with the original audio embedding. This fused embedding replaces the audio-only embedding for training the lip sync discriminator. Considering the Mel spectrum varies significantly across speakers and even across sentences from the same speaker, phonemes are a relatively stable feature that is consistent as long as the speech content remains the same, regardless of speaker or style. Thus, combining audio with phonemes leads to more accurate lip sync guidance and more stable lip motion synthesis. Denoting the phoneme embedding as $\mathit{p}$, the ultimate sync loss is formulated as:
\begin{equation}
    \mathcal{L}_{sync}^{'} = \frac{1}{N_i}\sum_{i}^{N_i}[\textnormal{-}y_i\log{S(a_i + {p_i}, v_i)} - (1-y_i)\log{(1-S(a_i + {p_i}, v_i))}]
\end{equation}
Once training stage 1 is done, the lip sync discriminator is fixed and serves as guidance for training stage 2. In this stage, the sync loss penalizes the mismatched motion of synthesized lips by comparing the video with the fused audio-phoneme reference, prompting the lip generator to produce more realistic, synchronized, and fluent lip image frames. 

\subsection{Stage 2: Lip Synthesis}

\textbf{Lip synthesis networks} Similar to previous studies~\citep{prajwal2020lip, wang2023seeing, gupta2023towards}, our lip generator uses a Seq2Seq network for reconstruction, comprising two audio and video encoders with an additional phoneme encoder and one decoder. Like the lip sync discriminator, the generator processes a triple tuple input, including an audio clip, a phoneme sequence with durations, and an image frame. The image frame is stacked on the RGB channels with two images from a video clip, one randomly selected as a full identity reference while the other with its lower half masked to predict the lip pose. The audio and visual elements are fed into a CNN-based encoder, while the phoneme sequence and duration are processed by the Transformer-based encoder described in Sec. \ref{SecStage1}, generating three embeddings. These embeddings are combined and passed to the CNN-based decoder, which generates the output layer by layer. Finally, the predicted face is separated from the stacked image and applied to the original video. The objective is to synthesize a facial image that closely resembles the original face but with the lip driven by the reference audio and phonemes. Given $\mathit{N_i}$ pairs of synthesized facial images $\mathit{v^{'}}$ and ground truth images $\mathit{v}$, we adopt L1 loss as the reconstruction loss between the synthesized and real facial images, which is calculated as:
%As in most prior studies, the lip generator in our work is based on a Seq2Seq network to conduct reconstruction, consisting of two audio and video encoders with an additional phoneme encoder and one decoder. Like the lip sync discriminator, the generator receives a similar triple tuple, including an audio clip, a phoneme series with duration for each, and an image frame stacked on the channels of RGB by two different images from a video clip, one stochastically picked as full identity reference while the other one's lower half masked to predict the lip pose. The acoustic and visual elements are given to the CNN-based encoder, while the phoneme sequence and duration are given to the Transformer-based encoder in Sec. \ref{SecStage1}, generating three embeddings, respectively, which are appropriately combined and fed into the CNN-based decoder to generate the final output layer by layer. Finally, the prediction is separated from the stacked image and pasted onto the original face when inferred. The goal of the generator is to synthesize face image as close to the original face as possible but possess a corresponding lip driven by the reference audio and the phonemes behind it. Given $\mathit{N_i}$ pairs of synthesized facial images $\mathit{v^{'}}$ and ground truth facial images $\mathit{v}$, we adopt L1 loss as the reconstruction loss between the synthesized and real facial images is formulated as:
\begin{equation}
    \mathcal{L}_{rec} = \frac{1}{N_i}\sum_{i}^{N_i}|v_i - {v_i}^{'}|
\end{equation}

In our lip synthesis networks, the synthesized images are augmented with noise through an adaptive diffusion model before being inputted to the visual discriminator, equivalent to the discriminator of GANs. This diffusion chain is a novel approach shown to improve training stability and efficiency of the GANs, which will be introduced in the following subsection.
%given to the visual discriminator, equivalent to the discriminator of GANs, are noise-augmented first by an adaptive diffusion model in the following subsection as a diffusion chain which is a novel approach proved to train GANs more efficiently. 

Let us denote $\mathit{D}$ as the visual discriminator in Fig. \ref{FigStage1}. 
The generator and discriminator losses caused by the visual discriminator are defined as:
\begin{equation}
    \mathcal{L}_{gen} = \frac{1}{N_i}\sum_{i}^{N_i}\textnormal{-}\log(1 - D(v^{'})) 
\end{equation}
\begin{equation}
    \mathcal{L}_{disc} = \frac{1}{N_i}\sum_{i}^{N_i}[\textnormal{-}\log{D(v)} - \log(1 - D(v^{'}))] 
\end{equation}
The generator loss $\mathcal{L}_{gen}$ propagates gradients back to improve the quality of synthesized facial images, while the discriminator loss $\mathcal{L}_{disc}$ strengthens the ability of discriminator to distinguish between synthesized and real facial images. Together, these losses drive the mutual reinforcement of the GANs.

\textbf{Optical flow consistency loss} The optical flow consistency loss is commonly used in stereo matching~\citep{lai2019bridging} and multi-view stereo tasks~\citep{furukawa2015multi}, comparing luminance consistency and motion smoothness between consecutive frames. Considering our task is to generate continuous image frames for fluent video output, we adopt the optical flow consistency loss as part of the total guidance of the generator to penalize anomalous motion variations among synthesized facial images. Unlike previous works such as Wav2Lip, which simply focus on audio-video consistency, our approach also ensures video inter-frame consistency. This is especially important when the original video features significant motion, with frequent changes in lip angle and pose.

To calculate this loss, we estimate the optical flow consistency between synthesized and real image sequences using a pre-trained RAFT model~\citep{teed2020raft}, applying L1 loss as the optical flow consistency metric. Let $\mathit{F(m, n)}$ represent the optical flow estimating function for two dynamic images, $\mathit{m}$ and $\mathit{n}$. Given $\mathit{N_i}$ pairs of synthesized facial images $\mathit{v^{'}}$ and ground truth images $\mathit{v}$, the optical flow consistency loss is defined as:

%The optical flow consistency loss is generally adopted in stereo matching and multiple view stereo tasks. It is calculated by comparing luminance consistency and motion smoothness between consecutive frames. Considering our task is to generate continuous image frames that demands high contiguity of each frame to produce a fluent video ultimately, we also adopt the optical flow consistency loss as part of total guidance of the generator to penalize anomalous motion variations among synthesized facial images, where most related work such as Wav2Lip overlooked and simply concentrated on the consistency between the audio and video rather than video itself. This is significantly useful when the original video is on a large range of motion with the angle and pose of lip changing frequently. Therefore, we estimate the optical flow consistency among each synthesized and real image sequences separately by a pre-trained RAFT model~\citep{teed2020raft} and adopt L1 loss as the optical flow consistency loss between synthesized and real image sequences. Let us denote $\mathit{C(m, n)}$ as the optical flow consistency calculating function where $\mathit{m}$ and $\mathit{n}$ are two estimated values of corresponding optical flow, given $\mathit{N_i}$ pairs of synthesised facial images $\mathit{v^{'}}$ and ground truth images $\mathit{v}$, the optical flow consistency loss is formulated as:
\begin{equation}
    \mathcal{L}_{cons} = \frac{1}{(N_i - 1)}\sum_{i=2}^{N_i}|F({v^{'}}_i, {v^{'}}_{i-1}) - F(v_i, v_{i-1})|
\end{equation}
Finally, the total loss for optimizing the lip synthesis networks combines all the aforementioned loss components and is formulated as follows:
%In summary, all the aforementioned losses about the generator are integrated to optimize the lip synthesis networks, which are formulated as:
\begin{equation}
    \mathcal{L}_{total} = \lambda_{sync}\cdot\mathcal{L}_{sync}^{'} + \lambda_{rec}\cdot\mathcal{L}_{rec} + \lambda_{gen}\cdot\mathcal{L}_{gen} + \lambda_{cons}\cdot\mathcal{L}_{cons}
\end{equation}
where $\lambda_{sync}$, $\lambda_{rec}$, $\lambda_{gen}$ and $\lambda_{cons}$ are scale factors that adjust the contributions of loss components.

\textbf{Adaptive diffusion model} %For stable training of GANs, injecting instance noise into the input of the discriminator is considered as a theoretically sound solution, which has not yet been carried out well in practice. Nevertheless, the method proposed in the work of Diffusion-GAN employs a Gaussian mixture distribution, defined over all the diffusion steps of a forward diffusion chain, to inject instance noise.
Inspired by Diffusion-GAN~\citep{wang2023diffusiongan}, which proposes a Gaussian mixture distribution over all diffusion steps in a forward length-adaptive diffusion chain to improve the stability and efficiency of GANs training, we integrate a similar technique into our framework. While maintaining the original GANs, we employ an additional diffusion model to noise-augment the facial images fed into the discriminator. This leads to enhanced training performance, as the generator benefits from its gradients backpropagating through the forward diffusion chain. The chain's length is adaptively adjusted by controlling the noise proportion added to both synthesized and real facial images, based on the discriminator's performance.

%maintain the original GANs as well but adopt an extra diffusion model as an effective approach to noise-augment the facial images inputting into the discriminator, leading to better training outcomes. The performance of the generator is positively affected by the process of its gradient backpropagating through the forward diffusion chain, whose length, in the meantime, is adaptively adjusted by controlling the proportion of noise added to the synthesized and real facial images according to the performance of the discriminator. 

The integration of adaptive diffusion model between the generator and discriminator will be demonstrated to accelerate convergence and stabilize the training process in Sec. \ref{SecAblation}, marking a successful practice of injecting instance noise in lip synthesis tasks. 

\section{Experiments}

\subsection{Experiments Settings}
\textbf{Dataset} We train our model using the LRS2 dataset~\citep{afouras2018deep} and evaluate it on unseen test sets of both the GRID~\citep{cooke2006audio} and LRS2 datasets. The GRID is a large multi-talker audiovisual sentence corpus whose video files have a resolution of 720$\times$576 and a frame rate of 25 fps. The audio from each video file is extracted with a maximum amplitude value of 1 and downsampled to 16 kHz. Sentences of GRID consist of a relatively fixed length of independent short words. The LRS2 is an open-world audio-visual speech recognition dataset whose video and extracted audio files have the same parameters of 25fps and 16kHz with GRID, respectively. Unlike GRID, sentences of LRS2 have more meaningful content of varying lengths, and the scenes are more diverse and irregular, making LRS2 more reflective of real-life scenarios.
%and the character forms and scenes are more varied and irregular, which is closer to reality.

\textbf{Metrics} 
We evaluate lip synchronization and the quality of synthesized images using widely used metrics such as FID~\citep{heusel2017gans}, SSIM~\citep{wang2004image}, LSE-D and LSE-C~\citep{prajwal2020lip}. LSE-D and LSE-C are calculated via a pre-trained sync net to measure the synchronization of lip movements, while FID and SSIM quantitatively assess image quality. In addition, we proposed a novel metric called Phoneme Error Rate (PER), which evaluates lip pose intelligibility and video fluency. PER is computed by comparing phonemes predicted from synthesized video with real phonemes extracted from audio, using a pre-trained lip-reading model AV-HuBERT~\citep{shilearning}. Unlike the Word Error Rate (WER) metric proposed by AV-HuBERT and adopted by TalkLip~\citep{wang2023seeing} in TFG, PER focuses directly on phonemes, avoiding the shortcomings of word-based evaluation, as the same phoneme sequence can represent multiple distinct words.

%comparisons of lip synchronization and quality of synthesized images of different methods are evaluated on the widely used metrics, including LSE-D, LSE-C~\citep{prajwal2020lip}, FID~\citep{heusel2017gans}, and SSIM~\citep{wang2004image}, where LSE-D and LSE-C are calculated via a pre-trained sync net to measure the synchronization of a series of lip motions while FID and SSIM assess the image quality quantitatively. In addition, we proposed a novel metric to conduct a comprehensive evaluation of the synthesized facial video, called Phoneme Error Rate (PER), measuring the lip pose intelligibility and video fluency. A pre-trained model of the lip-reading expert, Audio-Visual Hidden Unit BERT (AV-HuBERT)~\citep{shilearning}, is adopted to calculate the PER, which is specifically the proportion of edit distance between the phoneme characters transformed from the text predicted by the lip-reading expert from the synthesized video and the phoneme characters extracted from the real audio. Rather than compare the edit distance of texts such as the Word Error Rate (WER) proposed by TalkLip~\citep{wang2023seeing}, PER, which can be seen as an improved approach to traditional WER, is theoretically more precise and reliable for the evaluating task, for a same string of phoneme can share with a large number of distinct words which is an avoidable shortcoming when focusing on phonemes themselves. 

\textbf{Baselines} We compare our model against several SOTA lip synthesis models, including ATVGnet~\citep{chen2019hierarchical}, Wav2Lip~\citep{prajwal2020lip}, SadTalker~\citep{zhang2023sadtalker}, TalkLip~\citep{wang2023seeing}, and Diff2Lip~\citep{mukhopadhyay2024diff2lip}. ATVGnet is the first model to use an Attention-based Transformer Network (AT Network) and a Video Generator Network (VG Network) for generating talking face videos. Wav2Lip introduced the innovative use of a lip sync net in its reconstruction-based method. SadTalker generates videos by leveraging intermediate 3D Morphable Models (3DMM) and a 3D-aware face renderer. TalkLip builds upon Wav2Lip by integrating lip-reading loss and contrastive loss with guidance from a lip-reading expert. Diff2Lip is the latest SOTA model, adopting a diffusion model instead of the traditional Seq2Seq framework, achieving superior performance in lip synthesis.

%Primarily we compare our model with the following representative or SOTA models including ATVGnet~\citep{chen2019hierarchical}, Wav2Lip~\citep{prajwal2020lip}, SadTalker~\citep{zhang2023sadtalker}, TalkLip~\citep{wang2023seeing} and Diff2Lip~\citep{mukhopadhyay2024diff2lip}, using the aforementioned metrics. ATVGnet is the first work to adopt an Attention-based landmark Transformer Network (AT Network) and a Video Generator Network (VG Network) to produce talking faces videos. Wav2Lip is a representative work of reconstruction methods and innovatively introduces the lip sync net into its training process. SadTalker is a intermediate representation-based method that built a 3D Morphable Models (3DMM) and used a 3D-aware face renderer to generate videos. TalkLip makes full use of the guidance of a lip-reading expert, adding lip-reading loss and contrastive loss based on the former Wav2Lip model. Diff2Lip is one of the latest lip synthesis models and achieves SOTA performance for it adopts a diffusion model instead of the traditional Seq2Seq framework and reaches outstanding success of integration.

\textbf{Implementation Details} We have trained our models in environment configuration as follows: OS of Ubuntu20.04, CPU of AMD EPYC 9754 (18v CPU), GPU of RTX4090D (24GB) and RAM of 60GB. %Our model used to compare with other methods is integrated with all the proposed methods, 
Our model has been trained in stage 1 for 90k steps with a batch size of 40, and in stage 2 for 35k steps with the same batch size. To ensure fairness and rigor, we carry out the experiments for both our model and other approaches under the same setting and on a consistent range of the dataset.

\subsection{Experimental Results}
\begin{table}[!htp]
\caption{Quantitative performance comparisons of six different approaches on GRID and LRS2 datasets. PER is excluded from GRID due to the lack of semantic content in its sentences, making lip-reading predictions unreliable.}
%For every metric, each method is evaluated on the same group of samples, and the score is calculated by average. Unlike LRS2, PER is not applicable for GRID, for sentences in GRID lack semantics on the whole, which makes it hard for the lip-reading expert to predict the text content reasonably.} 
\label{TabExp1}
\renewcommand{\arraystretch}{1.4}
\resizebox{\linewidth}{!}{ 
\begin{tabular}{ccccccccccc}
\toprule[1pt]
\multicolumn{1}{c}{\multirow{2}{*}{Methods}} & \multicolumn{4}{c}{GRID}         & \multicolumn{5}{c}{LRS2}         \\ \cline{2-11}
\multicolumn{1}{c}{}                         & LSE-D$\downarrow$ & LSE-C$\uparrow$ & FID$\downarrow$ & SSIM (\%)$\uparrow$ & LSE-D$\downarrow$ & LSE-C$\uparrow$ & FID$\downarrow$ & SSIM (\%)$\uparrow$ & PER (\%)$\downarrow$ \\ 
\midrule[1pt]
Ground Truth                                 & 7.213 & 6.143 & 0.00 & 100.00 & 6.252 & 10.427 & 0.00 & 100.00 & 76.83 &     \\ \hdashline
ATVGnet                                      & 7.081 & 5.523 & 36.00 & 90.35 & 6.109 & 8.323 & 29.36 & 83.76 & 90.56 &     \\
Wav2Lip                                      & 6.352 & 6.627 & 26.71 & 96.10 & 5.487 & 11.516 & 28.80 & 91.93 & 77.92 &     \\
SadTalker                                    & 7.195 & 5.542 & $\textbf{20.08}$ & 87.80 & 5.524 & 9.792 & 98.50 & 55.59 & 73.91 &     \\
TalkLip                                      & 5.808 & $\textbf{7.534}$ & 35.38 & 95.85 & 5.755 & 10.561 & 22.71 & 92.64 & 47.31 &     \\
Diff2Lip                                      & $\textbf{5.710}$ & 6.903 & 33.70 & 95.37 & $\textbf{4.748}$ & 11.926 & 19.83 & $\textbf{94.54}$ & 82.93 &     \\ \hdashline
\textbf{FluentLip}                               & 6.258 & 6.790 & $\textbf{21.94}$ & $\textbf{96.25}$ & 5.018 & $\textbf{11.984}$ & $\textbf{16.93}$ & 93.31 & $\textbf{46.91}$ &     \\
\bottomrule[1pt]
\end{tabular}
}
\end{table}
\textbf{Quantitative results} The performance comparison of lip synchronization and visual quality of synthesized images of different approaches on the metrics mentioned above is shown in Tab. \ref{TabExp1} for both the GRID and LRS2 datasets. Guided by the diffusion model and optical flow consistency loss, FluentLip achieves near SOTA performance in terms of visual quality, realism and video fluency, with excellent synchronization. Our FluentLip attains top scores in FID, SSIM and PER, while also achieving competitive scores in LSE-D and LSE-C, which reflect lip synchronization.

FluentLip ranks second in LSE-D and first in LSE-C on the LRS2 dataset, highlighting the effectiveness of our phoneme-based multimodal learning strategy for improving synchronization. Moreover, FluentLip's standout performance in PER on the LRS2 dataset, especially when compared to models without the guidance of a lip-reading expert, underscores the model’s superior lip pose intelligibility and its accurate alignment between audio and lip movements. This further confirms the strength of our phoneme-based strategy. The performance on the GRID dataset, which is less varied in terms of background and speech content compared to LRS2, still shows FluentLip’s strengths in synchronization, as evidenced by its high LSE-D and LSE-C scores. FluentLip also demonstrates strong visual quality with leading FID and SSIM scores, reflecting its generalizability across unseen datasets.

%the significant advantage in PER on the LSR2 dataset demonstrates FluentLip's strong lip pose intelligibility in the generated videos and a more accurate correspondence between audio and mouth shapes, which accounts for the phoneme-based strategy as well. Compared to the LRS2 dataset, the GRID dataset is less varied in background and speech content. Still, FluentLip's advantages on synchronization reflected by LSE-D and LSE-C, and visual quality reflected by FID and SSIM carry over to the unseen GRID dataset, demonstrating FluentLip's strong generalizability.

\begin{figure}[!htp]
    \centering
    \includegraphics[width=1.0\linewidth]{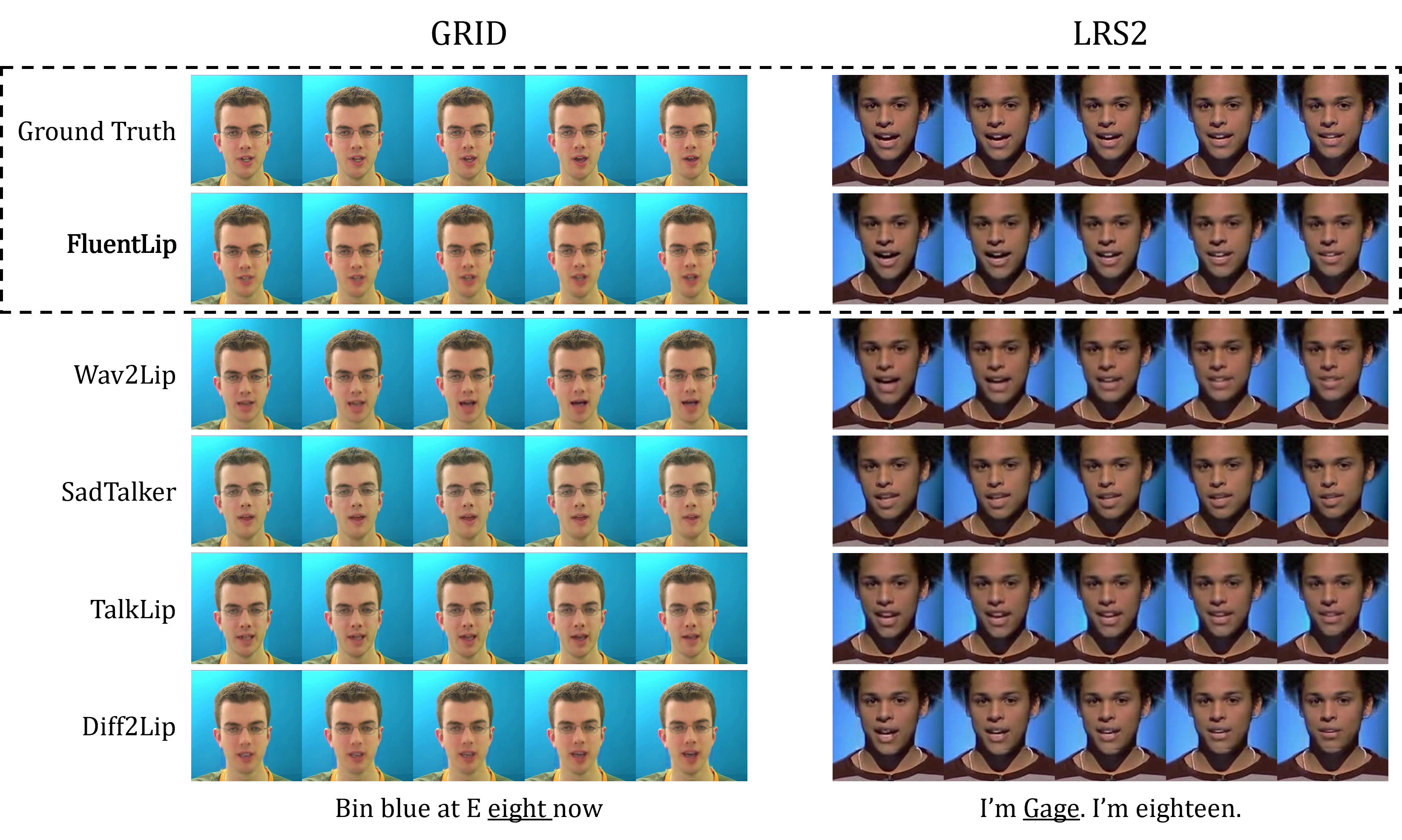}
    \caption{Qualitative comparison on five consecutive frames of the video from different approaches}
    \label{FigFaces}
\end{figure}

Both FID and SSIM are metrics that measure similarity between images, but in this case, we apply them to videos. FID evaluates the similarity in aspects such as visual quality, head motion trends, and lip poses, making it a comprehensive metric of video quality, synchronization and fluency. FluentLip’s FID is second only to SadTalker's on the GRID dataset and outperforms others on the LRS2 dataset, showing the highly competitive performance of FluentLip in balancing visual quality, synchronization, and fluency. SSIM, which directly measures image realism, places FluentLip ahead of Wav2Lip and TalkLip on the GRID dataset, and just slightly behind Diff2Lip on the LRS2 dataset, showing FluentLip's robustness in producing realistic video. It is worth noting that SadTalker, which generates facial animations from a single static image rather than a consecutive video, performs differently on the more static GRID dataset and on the more dynamic LRS2 dataset. As a result, SadTalker’s performance is optimized for datasets with fewer facial motions and expression changes, whereas FluentLip excels in handling more dynamic content like that found in LRS2.

\begin{figure}[!htp]
    \centering
    \includegraphics[width=0.95\linewidth]{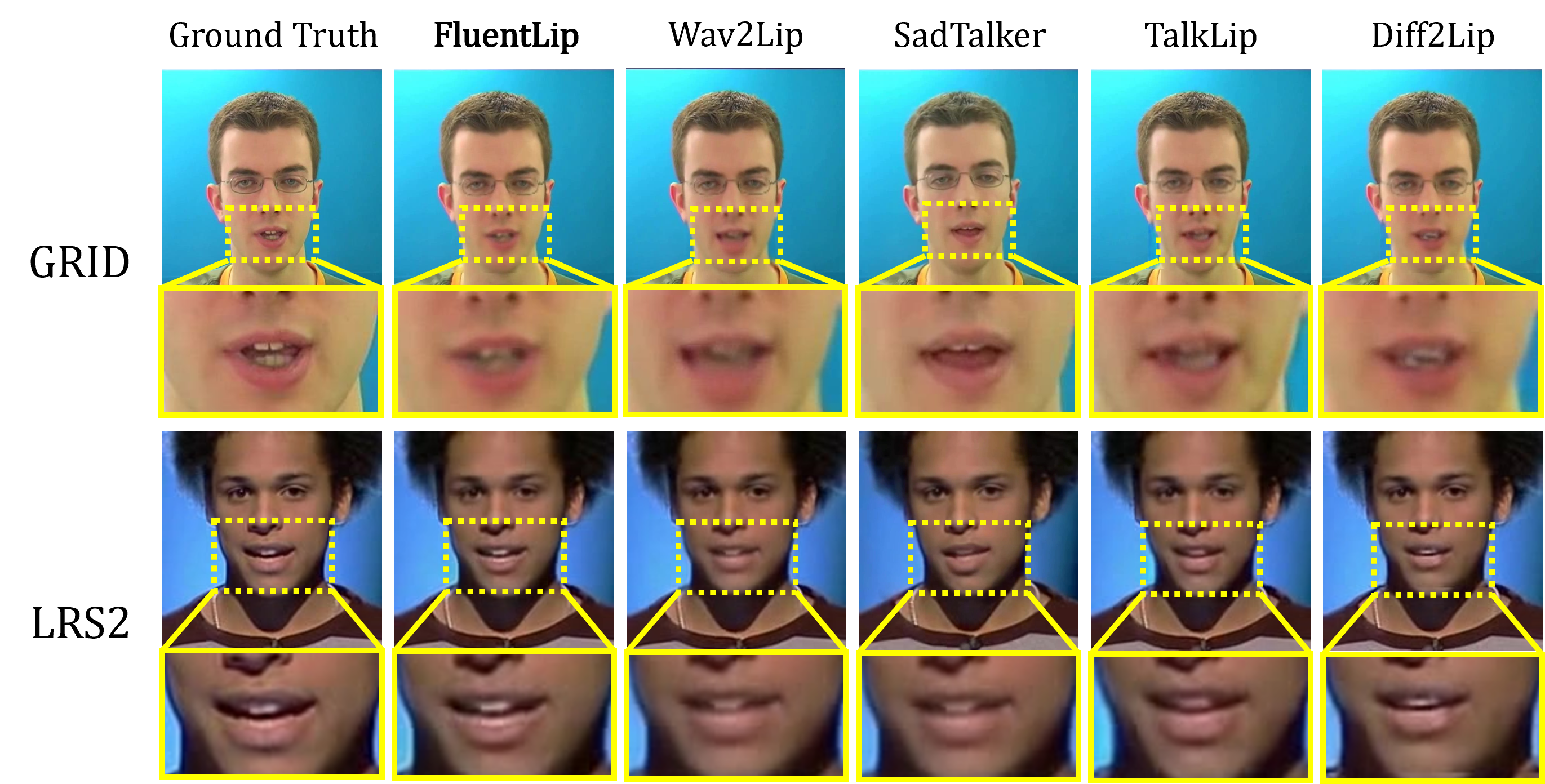}
    \caption{Single frame picked from Fig. \ref{FigFaces} and zoomed in on the lip region}
    \label{FigSingleFace}
\end{figure}

\textbf{Qualitative results} To qualitatively compare the videos generated by FluentLip with those generated by the other models, we present Fig. \ref{FigFaces}, which shows five consecutive frames from the videos generated by FluentLip and the different models using two arbitrarily selected videos and their corresponding audio from the test set as input. Specifically, the first row displays the ground truth video, and the second row shows the video frames generated by FluentLip, followed by the video frames generated by the other models in sequence. Moreover, we select and zoom in on a single lip pose from each image in Fig. \ref{FigFaces} to demonstrate differences in lip poses between FluentLip and the other models more closely, as shown in Fig. \ref{FigSingleFace}.

From Fig. \ref{FigFaces}, it is evident that FluentLip generates the most similar image frames to ground truth video regarding synchronization and smoothness. When compared to TalkLip and Diff2Lip, FluentLip generates highly consistent images with ground truth, without any abnormal color block in the video background. Furthermore, in comparison to Wav2Lip and Diff2Lip, FluentLip generates visible and significantly shaped teeth. Against SadTalker, FluentLip produces clear and natural faces with coherent and synchronized expressions and motions. Note that the five consecutive frames from SadTalker appear almost identical, suggesting that the use of 3D may lead to a static expression for the facial animation.

\subsection{Ablation Study}
\label{SecAblation}
To verify the effectiveness of each proposed key component, we have trained two variants of our FluentLip model under the following conditions: (1) without the integration of the optical flow consistency loss ($\textbf{FluentLip (w/o cons)}$), and (2) without the integration of the diffusion model ($\textbf{FluentLip (w/o diff)}$). For fair comparisons, FluentLip and its two variants have undergone the same training process in stage 1. In stage 2, they have been trained for 35,000 steps with a batch size of 40.

\begin{figure}[!htp]
    \centering
    \includegraphics[width=1\linewidth]{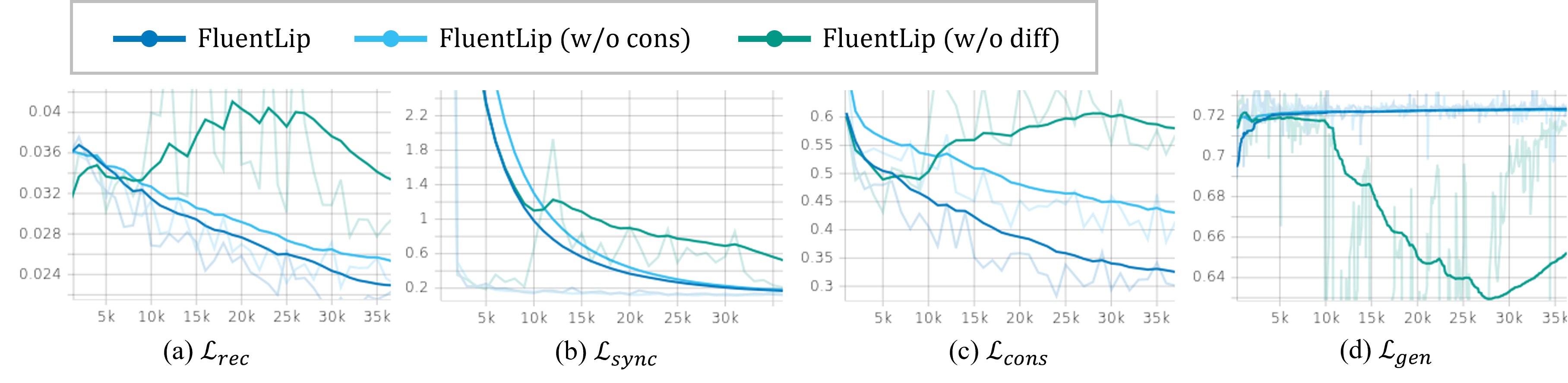}
    \caption{Variation of different losses of FluentLip and its two variants on the training of stage 2}
    \label{FigLoss}
\end{figure}
%Respectively, we have trained three versions of our models under the following conditions to verify the effectiveness of each proposal: with all the proposed methods integrated ($\textbf{FluentLip}$), without integration of the optical flow consistency flow ($\textbf{FluentLip (w/o cons)}$) and without integration of the diffusion model ($\textbf{FluentLip (w/o diff)}$). Given that the training of stage 1 is universal, three models above have been trained respectively on stage 2 for the same 35k steps with batch size of 40.

\textbf{Training results} To intuitively compare the performance of our FluentLip and its two variants during training, we select the variation of several crucial losses as shown in Fig. \ref{FigLoss}. 

First of all, regarding the diffusion model, FluentLip (w/o diff), which disables the diffusion chain during the GANs training, exhibits gradual mode collapse in the medium term. Despite losses getting down quickly at the beginning, the unstable training process leads to poor end results. This is evident in the downward and subsequent upward trend of losses (a), (b) and (c) in Fig. \ref{FigLoss}. This phenomenon is mainly caused by the repression of discriminator, as shown in Fig. \ref{FigLoss} (d). However, this issue is effectively mitigated by integrating the diffusion model, as demonstrated by FluentLip (w/o cons) and FluentLip.

When evaluating the impact of the optical flow consistency loss, FluentLip (w/o cons) consistently lags behind FluentLip, which utilizes the optical flow consistency loss. This clearly indicates that the optical flow consistency loss is beneficial for generating facial images with higher synchronization and visual quality. Models that adopt this loss function, such as FluentLip, achieve lower reconstruction and synchronization losses. 

Overall, these results provide strong evidence that all proposed key components positively influence both the training process and the final outcomes, confirming their effectiveness.

%the convergence rate of FluentLip (w/o cons) has been lagging thoroughly behind FluentLip, which used the optical flow consistency loss in contrast to the former. Accordingly, the optical flow consistency loss has proved helpful in generating facial images with higher synchronization and visual quality, for models with its adoption, such as FluentLip, possess a lower reconstruction loss as well as sync loss. All the above is an evident demonstration that all the proposed methods can have a positive impact on the training process and outcome, respectively, proving their usefulness.

\textbf{Quantitative results} We evaluated our FluentLip and two variants on both the GRID and LRS2 datasets using the same metrics as before. The comparisons across different metrics are presented in Tab. \ref{TabAblation}. As shown in the table, the quantitative performances of the three models generally align with the training results. Specifically, the overall metrics for FluentLip (w/o diff), FluentLip (w/o cons), and FluentLip exhibit a progressively superior trend, with FluentLip achieving the best results overall. Notably, the performance of FluentLip (w/o diff) varies significantly across different datasets, highlighting the instability of GANs, particularly concerning visual quality when the diffusion model is not utilized. The quantitative results, combined with the training findings, demonstrate that each of our proposed key components positively impacts the results, enhancing lip synchronization, visual quality as well as fluency.
%We take the same metrics above to evaluate our models in the dataset of both GRID and LRS2. The comparisons of different models on various metrics are shown in Tab. \ref{TabAblation}, respectively. As can be seen in the table, the quantitative performances of the three models are generally consistent with the training results, for the overall metrics of FluentLip (w/o diff), FluentLip (w/o cons) and FluentLip is mainly on a progressively superior trend, with FluentLip reaching the best outcome again. It's worth noting that the performance of FluentLip (w/o diff) varies widely on different dataset, representing the instability of GANs especially in terms of visual quality without the help of diffusion model. With the quantitative results above, together with the training results, we conclude that each of our proposals can have a positive impact on the results, leading to improvement in lip synchronization, visual quality, and most of all, overall fluency.

\begin{table}[t]
\caption{Quantitative performance comparisons of FluentLip and its two variants on GRID and LRS2 datasets.}% Results on the metrics are computed in the same way to Table \ref{TabExp1}.} % Methods with $\textbf{w.o.}$ are evaluated for the subsequent ablation study.}
\label{TabAblation}
\renewcommand{\arraystretch}{1.3}
\resizebox{\linewidth}{!}{
\begin{tabular}{ccccccccccc}
\toprule[1pt]
\multicolumn{1}{c}{\multirow{2}{*}{Methods}} & \multicolumn{4}{c}{GRID}         & \multicolumn{5}{c}{LRS2}         \\ \cline{2-11}
\multicolumn{1}{c}{}                         & LSE-D$\downarrow$ & LSE-C$\uparrow$ & FID$\downarrow$ & SSIM (\%)$\uparrow$ & LSE-D$\downarrow$ & LSE-C$\uparrow$ & FID$\downarrow$ & SSIM (\%)$\uparrow$ & PER (\%)$\downarrow$ \\ 
\midrule[1pt]
FluentLip                               & 6.258 & 6.790 & 21.94 & 96.25 & 5.018 & 11.984 & 16.93 & 93.31 & 46.91 &     \\
FluentLip (w/o cons)    & 6.825 & 6.485 & 23.13 & 96.22 & 5.629 & 11.206 & 24.60 & 90.54 & 66.25 &
\\
FluentLip (w/o diff)    & 7.594 & 5.917 & 82.48 & 94.37 & 5.048 & 11.928 & 25.81 & 91.20 & 68.37 &
\\
\bottomrule[1pt]
\end{tabular}
}
\end{table}

\section{Conclusion}
In this work, we have studied the challenges inherent in the talking face generation by proposing the FluentLip approach, which synthesizes facial videos with improved fluency and lip pose intelligibility. Unlike previous approaches that primarily focus on synchronization and visual quality, our FluentLip emphasizes lip intelligibility and video fluency by incorporating several novel components. We introduce optical flow consistency loss and utilize phonemes as input to enable multimodal learning, while also employing a diffusion model to stabilize the training of GANs. 

Extensive experiments demonstrate the effectiveness of the proposed FluentLip approach, showcasing highly competitive performances in lip synchronization and visual quality compared to five SOTA approaches from the literature. Notably, FluentLip outperforms these approaches in terms of fluency. In addition to these computational results, we conduct an in-depth analysis of the key components to shed light on their roles in the performance of the proposed approach.

\bibliography{iclr2025_conference}
\bibliographystyle{iclr2025_conference}

\end{document}

%% file: math_commands.tex
\usepackage{amsmath,amsfonts,bm}

\def\eqref#1{equation~\ref{#1}}
\def\1{\bm{1}}

\DeclareMathAlphabet{\mathsfit}{\encodingdefault}{\sfdefault}{m}{sl}
\SetMathAlphabet{\mathsfit}{bold}{\encodingdefault}{\sfdefault}{bx}{n}